\documentclass[10pt, a4paper]{article}

\usepackage{verbatim}
\usepackage[final]{lrec2026} 
\usepackage{xcolor}
\usepackage{multirow}
\usepackage{multicol}
\usepackage{graphicx}
\usepackage{amsmath}
\usepackage{float}

\title{What Triggers my Model? Contrastive Explanations Inform Gender Choices by Translation Models}

\name{Jani\c ca Hackenbuchner, Arda Tezcan, Joke Daems} 

\address{Language and Translation Technology Team, Ghent University\\
         \texttt{firstname.lastname@ugent.be}\\}

\abstract{
Interpretability can be implemented to understand decisions taken by (black box) models, such as neural machine translation (NMT) or large language models (LLMs). Yet, research in this area has been limited in relation to a manifested problem in these models: gender bias. In this work, we aim to move away from simply \textit{measuring} bias to \textit{exploring its origins}. Working with gender-ambiguous natural source data, this exploratory study examines which context, in the form of input tokens in the source sentence (EN), influences (or triggers) the NMT model’s choice of a certain gender inflection in the target languages (DE/ES). To analyse this, we compute saliency attribution based on contrastive translations. We first address the challenge of the lack of a scoring threshold and specifically examine different attribution levels of source words on the model’s gender decisions in the translation. We compare salient source words with human perceptions of gender and demonstrate a noticeable overlap between human perceptions and model attribution. Additionally, we provide a linguistic analysis of salient words. Our work showcases the relevance of \textit{understanding} model translation decisions in terms of gender, how this compares to human decisions and that this information should be leveraged to mitigate gender bias.
 \\ \newline  }

\begin{document}

\maketitleabstract

\section{Introduction}\label{intro}

Despite extensive efforts, gender bias continues to be exhibited by machine translation models, with no easy fix \citep{savoldi-etal-2025-decade}. Research on gender bias has primarily focused on coreference in unambiguous translation scenarios and has seldom been approached by means of interpretability. As interpretability can be a ``useful debugging tool for detecting bias in machine learning models'' \citep{molnar2025}, this paper tentatively explores the interplay of gender bias in machine translation (MT) and interpretability and explainability (XAI) to shine a light on \textit{how} a translation model handles gender, by \textit{explaining} the model's decisions.

Interpretability research in natural language processing (NLP) covers a variety of methodologies of assessing model behaviour, e.g. by focusing on the contributions of input tokens toward model predictions using feature attribution \citep{madsen-etal-2022}.
\citet[p.42202]{roscher-2020} describe interpretability as ``the mapping of an abstract concept [...] into a domain that the human can make sense of''. In this paper, we follow this understanding as a simplified form, considering a model as interpretable if its operation can be understood by humans \citep{biran-courtenay-2017}. We see interpretability as a tool to better understand (black box) model behaviour, allowing us to detect sources of gender bias in MT models and provide informed explanations for model decisions. 

Translating target referents, words that denote human referents (e.g. the word `reader'), from a notional or genderless language to a grammatical gender language (e.g. \textit{Leser} or \textit{Leserin}, in German) falls under word sense disambiguation because MT models have to translate ambiguous words semantically correctly \citep{rios-gonzalez-etal-2017}. In coreference scenarios, the MT model can fall back on the pronoun to disambiguate the gender of a target referent. In the absence of coreference, however, the MT cannot fall back to anaphora resolution, but instead tends to disambiguate gender based on stereotypes and occurrence in training data.

Previous work on implementing interpretability techniques to understand gender bias decisions in models is based on the extent to which models rely on coreferent gendered pronouns when translating a target referent (pronoun anaphora resolution) \citep{sarti-etal-2023-inseq,attanasio-etal-2023-tale,manna-etal-2025-paying}. These studies take coreferent pronouns as context to measure gender bias in (artificial) unambiguous sentences. It has not yet been examined how source words that have not previously been defined or are coreferent to a target referent affect a model's translation in terms of gender. From a human perspective, however, it has been shown that certain contextual cues influence (trigger) a human’s perception of gender, even in an ambiguous sentence context \citep{hackenbuchner-etal-2025-clin}. Furthermore, previous research lacks a defined scoring threshold as to which attribution level includes input tokens that are `influential enough' for the model’s gender decisions in the translation.

To fill this gap, this exploratory study is conducted on a small sample of English, natural gender-ambiguous data and, applying interpretability techniques, aims to understand which source words influence the NMT model’s choice of a certain gender inflection on target referents in the translation (English$\rightarrow$German/Spanish). We compare these results to human annotations, in a similar vein as \citet{yin-etal-2021} and \citet{sarti-etal-2024}.
To this end, we pose the following research questions: 
\begin{itemize}
    \item \textbf{[RQ1]} Can we effectively measure attribution level of source words on the MT model's gender decisions in the translation?
    \item \textbf{[RQ2]} Which (type of) source words are most salient for the gender translation of a target referent?
    \item \textbf{[RQ3]} To what extent do salient words for an MT model agree with the ones for humans?
\end{itemize}

This paper thus presents a novel experimental study\footnote{Data and code on GitHub: \url{https://github.com/jhacken/ContrastiveExplanations_GenderAttribution}.} on (i) interpretability methods for understanding MT choices on gender-ambiguous natural data, (ii) a computational linguistic analysis on the MT model's gender decisions, and (iii) a 
comparison of salient words and human annotations, in the vein of plausibility. With this study, we contribute to the limited research conducted on interpretability measures of model decisions in the translation of gender, particularly in natural, ambiguous scenarios.

\section{Related Research}\label{related}


\paragraph{Gender Bias, Ambiguity and Context} 
Gender bias is a fundamentally challenging task to solve due to a combination of factors \citep{vanmassenhove-2024}: (i) language models are trained on inherently biased data, where certain genders are under- or misrepresented, (ii) statistical and algorithmic bias introduced by the translation models exacerbate biases in outputs, (iii) societal stereotypes and norms surround gender roles in society, which, in turn, is embedded in training data, and (iv) languages are linguistically and grammatically different, with many requiring traditional grammatical gender markers, while others do not (e.g., `\textit{O} bir doktor' in Turkish is fully gender-neutral could be `He/She/They is/are a doctor' in English, while `\textit{She} is a doctor' in English requires a gendered pronoun, and `\textit{Sie} ist \textit{Ärztin}' in German requires both a gendered pronoun and a gendered referent). 

Early research on gender bias has shown that, based on biased training data, word embeddings can be gender-inflected, resulting in stereotypical translations (e.g. `doctor' being translated as male, and `nurse' being translated as female) \citep{caliskan-etal-2022,zhao-etal-2019,basta-etal-2019,Bolukbasi-etal-2016}. Since then, research has predominantly focused on assessing and demonstrating gender bias in translation for a combination of models and languages (\citealp{kotek-etal-2023,kocmi-etal-2020}, \textit{inter alia}), including on lesser researched languages (e.g., \citealp{gkovedarou_gender_2025,sewunetie-etal-2024,cho-etal-2019}), on constructing (handcrafted) challenge sets to evaluate and balance gender in translation (\citealp{pranav-etal-2025,savoldi-etal-2025-mgente,levy-etal-2021,bentivogli-etal-2020,stanovsky-etal-2019}, \textit{inter alia}) and on initiatives to mitigate gender bias, e.g., through domain-adaptation \citep{saunders-byrne-2020}, post-editing gender-fair translation \citep{lardelli_gender-fair_2024,nunziatini-diego-2024}, or by developing a gender-rewriting approach for gender-fair or neutralised translations \citep{savoldi-etal-2024-prompt,Piergentili:23-neutral,veloso-etal-2023,amrhein-etal-2023,vanmassenhove-etal-2021-neutral}.

Current research predominantly evaluates unambiguous gender-scenarios, where bias can be evaluated based on whether models correctly translate gender based on intra- or intersentential co-reference, where pronouns serve as contextual cues. At the same time, most research has been conducted on artificial, template-based sentences, for which reason large-scale natural, ambiguous data does not exist. Limited research has been conducted on (partially) gender-ambiguous data, where gender was predominantly measured with respect to a speaker, listener, plural referents or based on names \citep{savoldi-etal-2025-mgente,costa-jussa-etal-2023,saunders-olsen-2023,vanmassenhove-monti-2021,habash-etal-2019}. Working with fully gender-ambiguous source data, where gender is measured w.r.t. one specific referent has been evaluated on handcrafted data in \citet{hackenbuchner-etal-2025-genderous} and addressed on natural data in \citet{hackenbuchner-etal-2025-clin}.

Previous papers frequently refer to coreference anaphora resolution in MT as `context'. In this paper, however, we argue that coreference between a target referent noun and a pronoun does not offer sufficient contextual information, being a case of (necessary) linguistic agreement. In contrast, we define `context' to be all words in a sentence that provide semantically relevant information for a human or a model to disambiguate a term with. 
As an example, in the sentence ``The patient went into labour'', the contextual information of ``[going] into labour'' disambiguates the gender of the `patient' as very likely female. In this paper, we focus on `context' as exemplified in the last example, which can be more subtle, lacking a clear resolution for the gender of the target referent in question. The data and phenomena analysed in this work will be outlined in more detail in Section \ref{methodology}.

\paragraph{Interpretability and Gender Bias} Research on using interpretability as a tool to understand and inform solutions for gender bias in a translation context has been examined sporadically. Some papers used interpretability as a method to specifically assess gender bias \citep{vamvas-sennrich-2021,costa-jussa-etal-2022,attanasio-etal-2023-tale,manna-etal-2025-paying,conti-etal-2025-unheard,conti-etal-2025}, while others focussed on model interpretability in general, with only a side note on gender bias \citep{voita-etal-2018,yin-etal-2021,sarti-etal-2023-inseq}. Informed by their results, \citet[p.425]{sarti-etal-2023-inseq} support ``the validity of contrastive approaches in uncovering gender bias''. Similarly, \citet{attanasio-etal-2023-tale} ``prove interpretability as a valuable tool for studying and mitigating bias in language models'' (p.3997). These claims confirm this study's choice of interpretability as a means to examine and understand gender bias in translation models.

In these works, gender bias has either been examined by means of attention-based techniques, where attention weights are leveraged as a window into model decision-making \citep{costa-jussa-etal-2022,manna-etal-2025-paying}, through saliency attribution techniques to understand which input features are responsible for a given output decision \citep{sarti-etal-2023-inseq,attanasio-etal-2023-tale}, or specifically through contrastive explanations, to compare model's output behaviour based on minimally different inputs \citep{vamvas-sennrich-2021,yin-neubig-2022-interpreting}, which has recently also been examined in speech translation \citep{conti-etal-2025-unheard,conti-etal-2025}. In this paper, we compute \textit{input saliency scores}, which have been argued to be best suited for XAI because they ``reveal why one particular model prediction was made in terms of how relevant each input word was to that prediction'' and do so by taking the ``entire computation path into account, all the way from the input word embeddings to the target output prediction value'' \citep[p.152]{bastings-filippova-2020}.

As interpretability is a `fuzzy concept' \citep{molnar2025}, there is no `one' specific threshold to measure attribution scores. Previous research establishes different thresholds for measuring whether computed scores and tokens are `influential enough'. \citet{manna-etal-2025-paying} consider attention heads that exceed a defined baseline based on average sentence length by a ``notable margin''. \citet{attanasio-etal-2023-tale} focus on a ``relative difference'' in scores between correct and wrong translations. \citet{sarti-etal-2023-inseq} correlate (w. Kendall's correlation) the different metrics they applied with how much the gender distribution deviates from an equal distribution. In this paper, we address the challenge of a lacking scoring threshold and specifically examine different attribution levels of source words on the model’s gender decisions in the translation in comparison to human annotations.

\paragraph{Contrastive Explanations} Contrastive Explanations demonstrate which input tokens lead a model to predict one output instead of another (why is X predicted instead of Y?) \citep{yin-neubig-2022-interpreting}. Contrastive explanations have been shown to outperform non-contrastive ones, particularly in longer-range context, and can help explain model predictions more accurately and clearly to a human observer \citep{yin-neubig-2022-interpreting}. They have successfully been studied both for gender \citep{vamvas-sennrich-2021,sarti-etal-2023-inseq} and for gender-unrelated topics \citep{sennrich-2017,rios-gonzalez-etal-2017,yin-neubig-2022-interpreting}. \citet{vamvas-sennrich-2021} propose contrastive conditioning to disambiguate gender bias in MT, for which they pair evaluated translations with contrastive source sequences that are slightly modified to provide a stronger disambiguation cue. Leveraging contrastive feature attribution in speech translation, \citet{conti-etal-2025-unheard,conti-etal-2025} reveal that to accurately translate gender, the model relies on first-person pronouns to link gendered terms back to the speaker. \citet{yin-neubig-2022-interpreting}, interestingly, interpreted language models using contrastive explanations and, even though their focus did not lie on detecting or assessing gender bias, they show that when a model generates a wrongly-gendered pronoun, the model tended to be influenced by proper nouns and pronouns of a different gender in the input. If a model correctly chooses gender for pronouns, nouns and determiners, it was often influenced by source gendered proper nouns and pronouns.

\paragraph{Plausibility} Plausibility, also referred to as `human-interpretability' \citep{lage-etal-2019}, refers to the alignment between model rationales and salient source words (input tokens) identified by human annotators \citep{jacovi-goldberg-2020}. Plausibility measures ``how convincing the interpretation is to humans'' \citep[p.4199]{jacovi-goldberg-2020} and has previously been explored in relation to gender bias in MT in \citet{sarti-etal-2024}. This concept is closely related to the methodology conducted in the explorative study presented in this paper.

\section{Methodology}\label{methodology}

\begin{figure*}[!t]
\centering
    \includegraphics[width=0.9\linewidth]{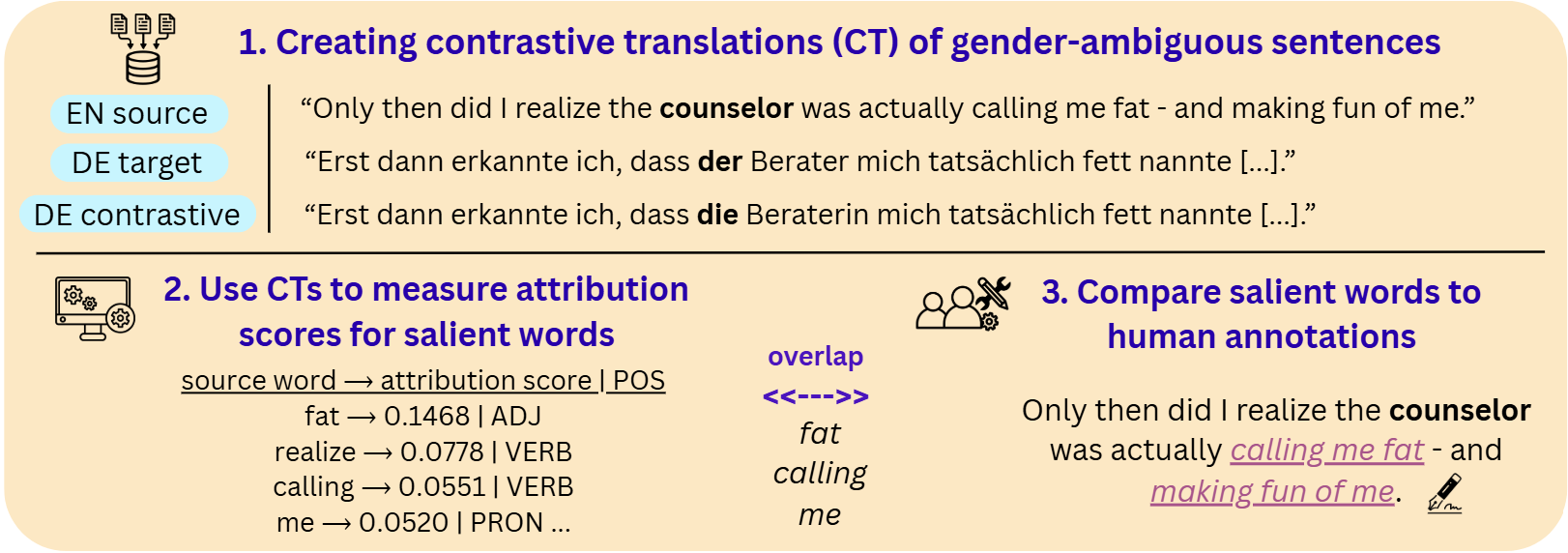} 
    \caption{Methodological outline of (i) creating contrastive translations from gender-ambiguous source sentences to (ii) computing attribution scores of input tokens to (iii) comparing the most salient source words to human annotations.}
    \label{fig:fig1}
\end{figure*}

\paragraph{Data} The English source data used for this research is taken from \citet{hackenbuchner-etal-2025-clin}, which comprises a set of manually-filtered natural, gender-ambiguous sentences that refers to a singular target referent (e.g. `writer'), as in ``For someone who is normally a business \textit{writer}, that is very valuable.'' The sample data consists of 60 sentences and 48 unique target referents (e.g., `writer', `poet'), with an average of 19.67 words per sentence and a standard deviation of 8.54. Unfortunately, there exists no larger sample of gender-ambiguous natural data and human annotations, on which the conclusions of this study could be based.

Starting from this English (EN) source data, we (i) translated the EN sentences into German (DE) and Spanish (ES) using \texttt{OPUS-MT}\footnote{\url{https://huggingface.co/Helsinki-NLP/opus-mt-en-de}} \citep{tiedemann2020}, and (ii) based on these original output translations, we created contrastive gender translations\footnote{The contrastive translations were created by the main author, who has (near-)native competence in the target languages and holds an MA degree in Specialised Translation.}, meaning that we grammatically adapted the translation to agree with the opposite gender in Spanish or German (exemplified in Figure \ref{fig:fig1}). We thus had 60 EN$\rightarrow$DE/ES original translation pairs and 60 EN$\rightarrow$DE/ES contrastive translation pairs, respectively.

The data in \citet{hackenbuchner-etal-2025-clin} includes human annotations by 20 annotators per sentence (of male, female and non-binary genders), indicating which source words affect their \textit{perception of gender} with respect to the target referent. The annotators followed a bottom-up approach, where they annotated any (or no) words that affected their individual perception of gender with respect to the target referent. The instructions were that for a specific target referent, they were asked to mark all individual words that they believed (could) affect the gender of the person in question. An example screenshot of the instructions is provided in Appendix \ref{app:annotation_instructions}. They show that (ambiguous) context has a demonstrable impact on human gender perceptions, which had initially been demonstrated in \citet{hackenbuchner-etal-2024}.

\paragraph{Saliency Attribution and Pre-Processing} We are interested in source words in the sentence context that contribute the most to a certain gender translation, i.e. which source words are \textit{most salient} for an MT model. Using the \texttt{OPUS-MT} model via the inseq toolkit \citep{sarti-etal-2023-inseq}\footnote{\url{https://github.com/inseq-team/inseq}} \texttt{version 0.6.0}, we compute saliency attribution on two contrastive outputs: (i) the original MT translation and (ii) a manually-created translation contrasting in terms of gender (a foil). 
Based on the work and formulations by \citet{yin-neubig-2022-interpreting}\footnote{Section 3 in \citet{yin-neubig-2022-interpreting} provides formulations for the gradient norm, how saliency scores are obtained and how this is extended to the contrastive gradient norm.}, saliency scores are computed based on the norm of the gradient of the model prediction with respect to the input. The default applied here takes the L2 norm to aggregate gradient vectors and takes the probability of the next word. This extends to the contrastive gradient norm, which measures how much an input token influences the model to increase the probability of the next token in the sequence while decreasing the probability of the foil token.

As exemplified in Figure \ref{fig:fig1} step 1, we take the original model translation [DE target] and contrast it to a gender-opposite translation [DE contrastive]. We analyse which source sentences [EN source] lead to a higher probability of being translated into masculine (e.g., DE: \textit{Berater}, original MT output) or into feminine (e.g., DE: \textit{Beraterin}, manually contrastive translation, i.e. foil), and specifically which input tokens affect this probability. Applying this contrastive explanation methodology, we compute saliency scores (depicted in Figure \ref{fig:fig1}, step 2) for the MT model's choice of translation for the first (original) target referent in contrast to the alternative (gender-opposite) target referent (foil). We focus on source text saliency (i.e. saliency of the tokens in the source text) as we want to measure the effect of input tokens on translation, and we compare these to human annotations that were conducted on English source data (depicted in Figure \ref{fig:fig1}, step 3). Given the nature of transformer-based NMT models, in which source tokens as well as previously generated target tokens influence the translations of the following words, this analysis also provides saliency scores for such previously generated target tokens, preceeding the token where the alternative gender translations are analysed. To exclude the impact of the attributions of the previously generated target tokens in the MT output from this analysis, we left-align tokens and analyse the first different token in the MT output (compared to the alternative translation with opposing gender) [e.g., \textit{der} $\rightarrow$ \textit{die}], while ensuring that the previously generated tokens remain the same in both translations. 

\begin{figure}
\centering
  \includegraphics[width=\columnwidth]{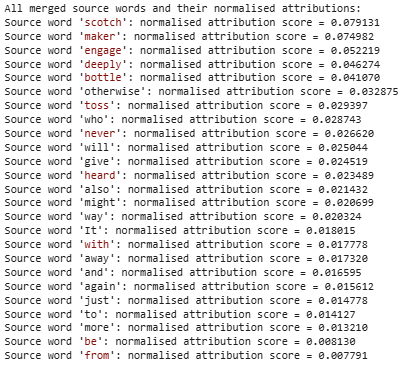}
  \caption {Example depiction of pre-processed source words and their normalised attribution scores, from highest to lowest. Words in red overlap with human annotations.}
  \label{fig:example_sent23}
\end{figure}

Prior to our analyses, we performed L1 normalisation on the source token attributions so that the attribution scores sum to 1, followed by four pre-processing steps: (1) we remove the target referent token [e.g., \textit{counselor}] to focus on contextual cues, (2) we remove end-of-sentence tokens </s> from our analysis, as we are interested in mostly \textit{content} words, (3) we remove punctuation marks and a limited set of stopwords, such as articles and determiners ("", a, an, the, this, that, these, those), for the same reasoning, and (4) we merge sub-word tokens and add their individual attribution scores, as was done in \citet{manna-etal-2025-paying}.

After the preprocessing steps, we implement the following four approaches to analyse saliency of input tokens from different perspectives and compare selected input tokens with human annotations for saliency. 
These approaches can be exemplified by looking at Figure \ref{fig:example_sent23}, which shows source words and their normalised attribution scores, from highest to lowest, for the following sentence: ``It will also give the scotch maker a way to engage more deeply with a \textit{consumer} who might otherwise just toss that bottle away and never be heard from again'', where the target referent is \textit{consumer}.

\paragraph{Approaches for Effectively Measuring Attribution Levels [RQ1]} We leverage the following four approaches to measure attribution level:

\begin{enumerate}
  \item We retrieve the top X\% of source words, sorted by their attribution score from highest to lowest, where we assessed an overall range of top 5--25\%, in 5\% increments. In Figure \ref{fig:example_sent23}, if we would take the e.g. top 10\% of all source words, our threshold would include \textit{scotch} [score: 0.0791] and \textit{maker} [score: 0.0750].
  \item We retrieve only the top word, i.e. the word with the overall highest attribution score. With this method, our threshold would only include the word \textit{scotch} [score: 0.0791] in Figure \ref{fig:example_sent23}.
  \item We retrieve individual words whose attribution scores exceed a certain threshold, where we assessed a range of 0.01--0.10 score per word. In Figure \ref{fig:example_sent23}, if we would take all source words with a minimum score of e.g. 0.06, our threshold would include \textit{scotch} [score: 0.0791] and \textit{maker} [score: 0.0750].
  \item We first calculate the total attribution score for a specific sentence\footnote{After removing the target referent and stopwords, the attribution scores of the remaining words within a sentence no longer sum to 1.} and then identify the minimum subset of input words whose cumulative attributions scores reach at least X\% of the total attribution scores combined. We assess thresholds of 5--50\%, in 5\% increments. In Figure 2, the e.g. top 20\% of summed attribution scores would include \textit{scotch} [score: 0.0791], \textit{maker} [score: 0.0750] and \textit{engage} [score: 0.0522].
\end{enumerate}

\paragraph{Model-Human Overlap [RQ3]} We are interested in how many of the source words considered salient by the MT model were also annotated by the annotators. To do so, we measure the overlap between salient source words and human annotations, taken from \citet{hackenbuchner-etal-2025-clin}, which we calculate as precision:

\[
\text{Precision} = \frac{|\text{Attribution} \cap \text{Annotation}|}{|\text{Attribution}|}
\]

We calculate micro precision by dividing the sum of all overlaps by the total number of salient words. This approach avoids extreme values obtained from shorter sentences. In comparison, we show macro precision scores (arithmetic mean) in the Appendix, Table \ref{tab_app:approaches}. We take all annotations into consideration for the comparison with salient words (excluding the specific stopwords we removed) from the MT model. However, it has previously been shown that human perceptions of gender vary extremely among annotators, with an overall `fair' inter-annotator agreement (IAA) ($\kappa$~=~0.364, Fleiss' Kappa \citep{fleiss}) \citep{hackenbuchner-etal-2024,hackenbuchner-etal-2025-clin}. 
To account for this variability, we also perform a comparative analysis restricted to words annotated by at least two annotators, i.e. cases where there was explicit agreement among the annotations.

\paragraph{Linguistic Analysis [RQ2]} We conduct a linguistic analysis of salient words to better understand which types of words in context influence a model's translation decision in terms of gender. To analyse parts-of-speech (POS) labels of salient words, we used the SpaCy toolkit\footnote{\url{https://spacy.io/}}, which performs POS tagging based on the Universal Dependencies annotation scheme\footnote{\url{https://universaldependencies.org/}}. Secondly, we compute the dependency distance between each salient word and the target referent, defined as the number of syntactic links (edges) separating the two nodes in a given dependency tree. 
We compare the POS label distribution of salient words and human-annotations. We also compare the distance of the salient words to the target referent both for the case of MT and human annotations. Finally, we examine the extent to which the model and humans are influenced by the same type of word (POS category) and by the same grammatical structure (dependency trees).

\section{Results}\label{results}

\texttt{OPUS MT} translated the majority of gender-ambiguous referents (and other agreeing words, e.g. pronouns) into masculine (88.3\% for EN$\rightarrow$DE and 83.3\% for ES), while into feminine a mere 11.6\% for EN$\rightarrow$DE and 13.3\% for ES. 3.3\% for EN$\rightarrow$ES were translated as neutral (e.g. \textit{amante}). In contrast, human annotations (majority taken) annotated target referents in 33.3\% of sentences as feminine. 
The following analysis is based on the source words that respectively influenced human gender perceptions (annotations) or MT choice (saliency).

\begin{table}
\centering
\begin{tabular}{p{0.05\linewidth}c|c|c|c|c}
\multicolumn{2}{l}{\textbf{Approach:}} & \textbf{1} & \textbf{2} & \textbf{3} & \textbf{4} \\
\hline
\textbf{DE} & $P_{mi}$  & .817 & .817 & .791 & .817 \\
\hline
\textbf{ES} & $P_{mi}$ & .879 & .879 & .831 & .897 \\
\hline
\hline
\textbf{Avg} & $P_{mi}$ & .848 & .848 & .811 & \textbf{.851} \\
                                                        
\end{tabular}
\caption{\label{tab:approaches}
    Highest model-human overlap achieved per approach (1-4) taken, for DE, ES and an overall average. $P_{mi}$ denotes micro precision.}
\end{table}

\subsection{Leveraging Approaches to Effectively Measure Attribution Levels of Source Words [RQ1, RQ3]} 
For the four implemented approaches to test different attribution levels of salient input words, we first analyse the thresholds leading to the highest overlaps with human annotations (i.e. highest precision scores), as shown in Table \ref{tab:approaches}:
\begin{enumerate}
\item Approach 1: The highest overlap between salient words and human annotations occurs for an overall top \textbf{5\%} of all source words, with 0.817 for EN$\rightarrow$DE and 0.879 for EN$\rightarrow$ES.
  \item Approach 2: For the \textbf{top one} most salient word, the overlap between salient words and human annotations is equally 0.817 for EN$\rightarrow$DE and 0.879 for EN$\rightarrow$ES.
  \item Approach 3: The highest overlap between salient words and human annotations occurs at an individual minimum score of \textbf{0.1}, with 0.791 for EN$\rightarrow$DE and 0.831 for EN$\rightarrow$ES.
  \item Approach 4: The highest overlap between salient words and human annotations occurs at the relative top \textbf{15\%} of attribution scores with 0.897 for EN$\rightarrow$ES and with 0.817 for EN$\rightarrow$DE at either top  5\%, 10\% and 20\%.
\end{enumerate}

We can make a number of observations from Table \ref{tab:approaches}, and find answers to RQ1 and RQ3: 1. all perspectives point to high overlaps with human annotations, 2. the lowest overlap occurs when we retrieve tokens above a fixed attribution score threshold (approach 3), 3. the highest overlap occurs for the top X\% of attribution scores relative to the sum of each individual sentence (approach 4), and 4. model-human overlap is highest when only a top few salient words are considered (e.g., top 5\%, top word, relative top 15\%).

\begin{figure}[t]
\centering
  \includegraphics[width=0.9\columnwidth]{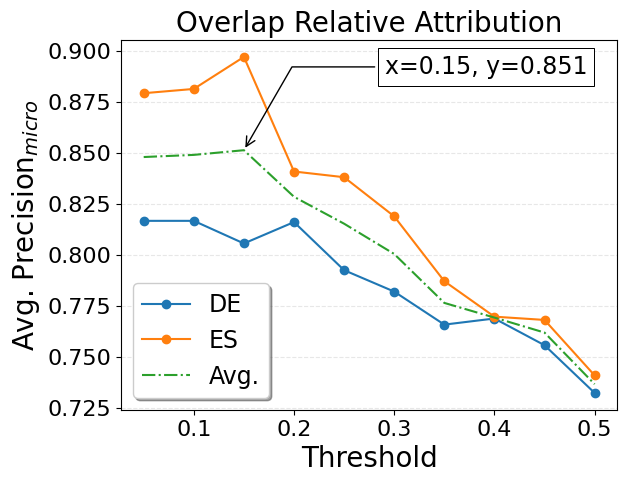}
  \caption {Comparative scoring of overlapping source words for Approach 4, depicting micro-averaged precision scores for EN$\rightarrow$DE/ES and their average. The x-axis should be interpreted in percent (e.g., 10\%).}
  \label{fig:approach4}
\end{figure}

As the average across EN$\rightarrow$DE/ES scores yields the highest overall overlap of 0.851 for the relative top 15\% (approach 4),  we focus our continued analysis on this approach. All micro-averaged precision scores of approach 4 for EN$\rightarrow$DE/ES are visualised in Figure \ref{fig:approach4}. A complete overview of all results of the different approaches (as well as macro precision scores) can be found in the Appendix, Figure \ref{fig_app:approach4} and Table \ref{tab_app:approaches}.

\begin{figure}[t]
\centering
  \includegraphics[width=\columnwidth]{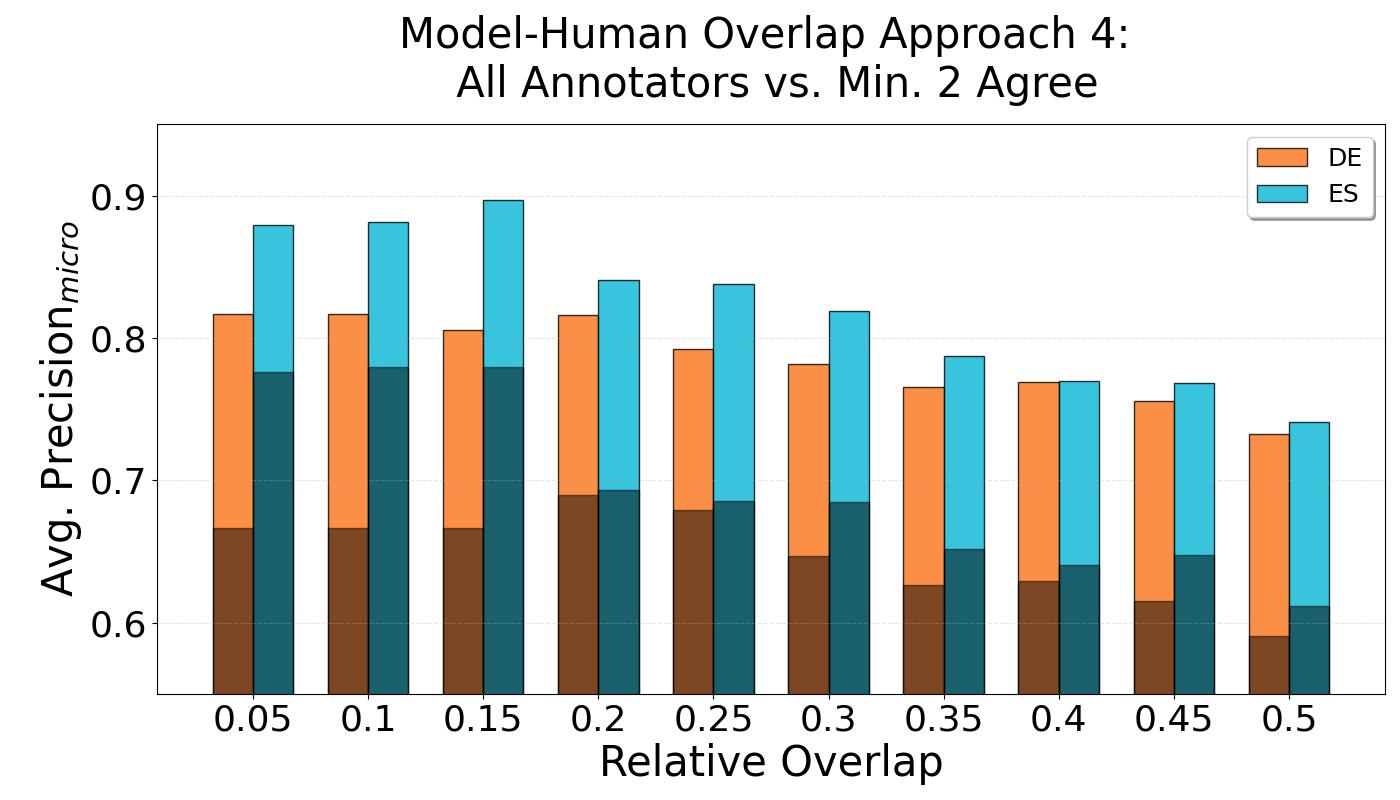}
  \caption {Approach 4: Comparison between micro-averaged precision scores for two annotation sets, all annotations vs.\ annotations where at least two annotators agree (i.e.\ source words annotated by at least two annotators), represented by the lower dark shaded part of the bars.}
  \label{fig:approach4_annotator_comparison}
\end{figure}

For our evaluation, all unique annotations were compared to salient source words. In comparison, we also tested a model-human overlap between annotated words where at least two annotators agreed. In other words, we excluded words that were annotated by only a single person. For approach 4, the comparative overlapping values achieved are depicted in Figure \ref{fig:approach4_annotator_comparison}. When only taking words into account that at least two annotators annotated, the highest overlap (i.e. micro precision) with the MT model achieved is 0.78 for ES and 0.69 for DE. This continues to be a noticeable overlap, but significantly lower than when comparing all annotations.
These results are not surprising: only taking source words into consideration that have been annotated by at least two annotators decreases the total number of source words available for comparison with the salient words indentified for the MT model. Some source words did not lead to an agreement among annotators (as they had only been annotated once) but are, in contrast, considered salient for the model's translation. As a result, this reduction in the number of source words can be considered as the main reason that leads to a lower number of overlaps.  
In the further linguistic analysis of this paper, we focus on overlapping source words that were annotated by any annotator at least once (result from Figure \ref{fig:approach4}).

\subsection{Linguistic Analysis of Salient Words [RQ2]}

The following paragraphs provide an overview of what type of words are considered salient by the model tested in the directions EN$\rightarrow$DE and EN$\rightarrow$ES, and how these words are syntactically linked to the referent. Both languages reveal similar patterns. Overall, of 1055 words (which excludes determiners and punctuation marks), DE and ES considered salient 72 and 67 words, respectively. Of these, the same $\sim$50\% of words are found to be salient for both $\rightarrow$DE and $\rightarrow$ES.

\paragraph{Parts-of-Speech}

\begin{figure}[t]
\centering
  \includegraphics[width=\columnwidth]{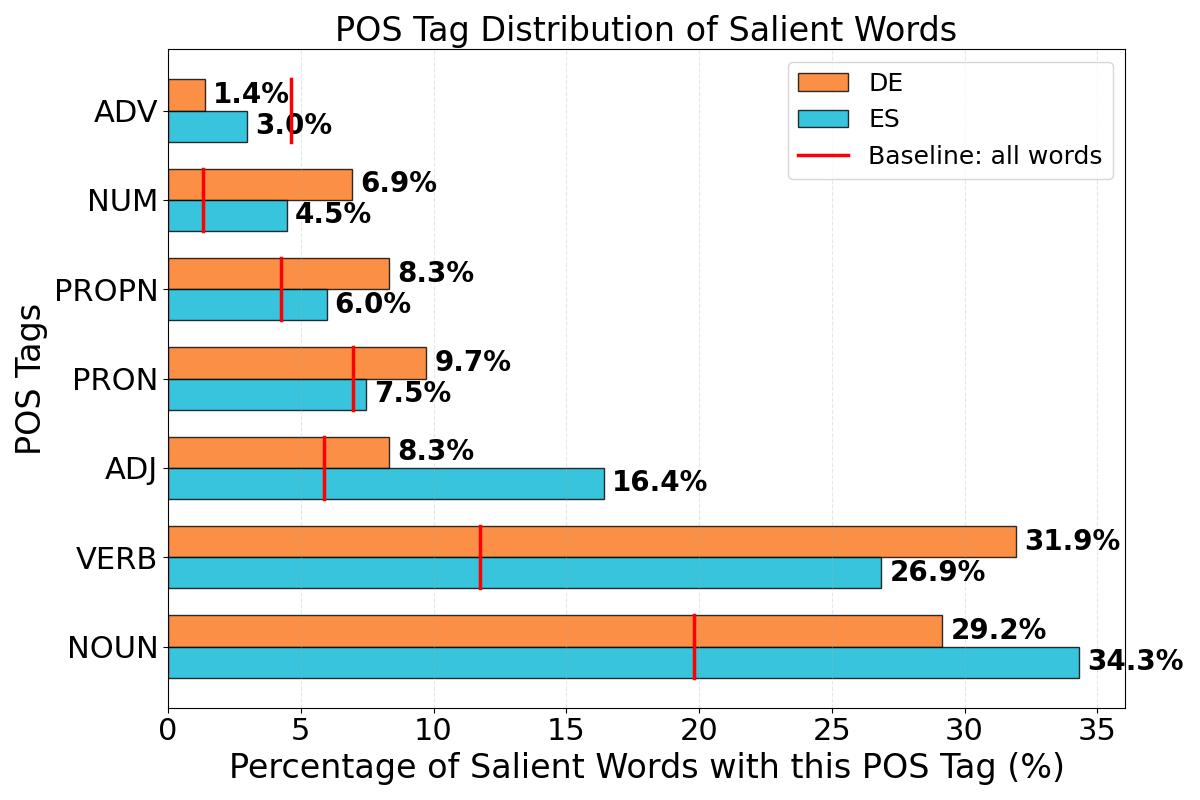} 
  \caption {Relative POS of salient words (Approach 4), in comparison to the distribution of POS tags in the source data.}
  \label{fig:POS}
\end{figure}

Figure \ref{fig:POS} shows the POS labels of all salient words for both EN$\rightarrow$DE and EN$\rightarrow$ES. The general trends of POS categories for context words considered salient are relatively similar for both language pairs. Nouns were the top-most annotated POS category, comprising 34.3\% and 29.2\% of all salient words for ES and DE, respectively. The second most annotated POS category is verbs, comprising 26.9\% and 31.9\% of all salient words for ES and DE, respectively. These two POS categories stand out among the others. They were followed, to a lesser extent, by adjectives, pronouns, proper nouns, and numbers. The general trend of these salient POS categories follows the relative occurrence of the overall POS frequency of all words, represented by the red vertical lines in Figure \ref{fig:POS}. However, it becomes clear that both nouns and, particularly, verbs (and adjectives for ES) are considered salient considerably more frequently than their occurrence in the overall data. To a lesser extent, this can similarly be seen for numbers, proper nouns, and pronouns.

In comparison, in \citet[p. 299]{hackenbuchner-etal-2025-clin} it has been shown that the context words that most affected human perceptions of gender had POS labels of (in steady, decreasing order) proper nouns, adjectives, nouns, verbs, pronouns, and adverbs. Despite the fact that both scenarios share similar dominant POS categories for context words, namely (proper) nouns, verbs, adjectives, and pronouns, the overall frequency distributions differ. In particular, the MT model predominantly marks nouns and verbs as salient source words, whereas humans highlight a different distribution of word types, indicating that the model and humans are influenced by different types of words.

\begin{figure}[t]
\centering
  \includegraphics[width=\columnwidth]{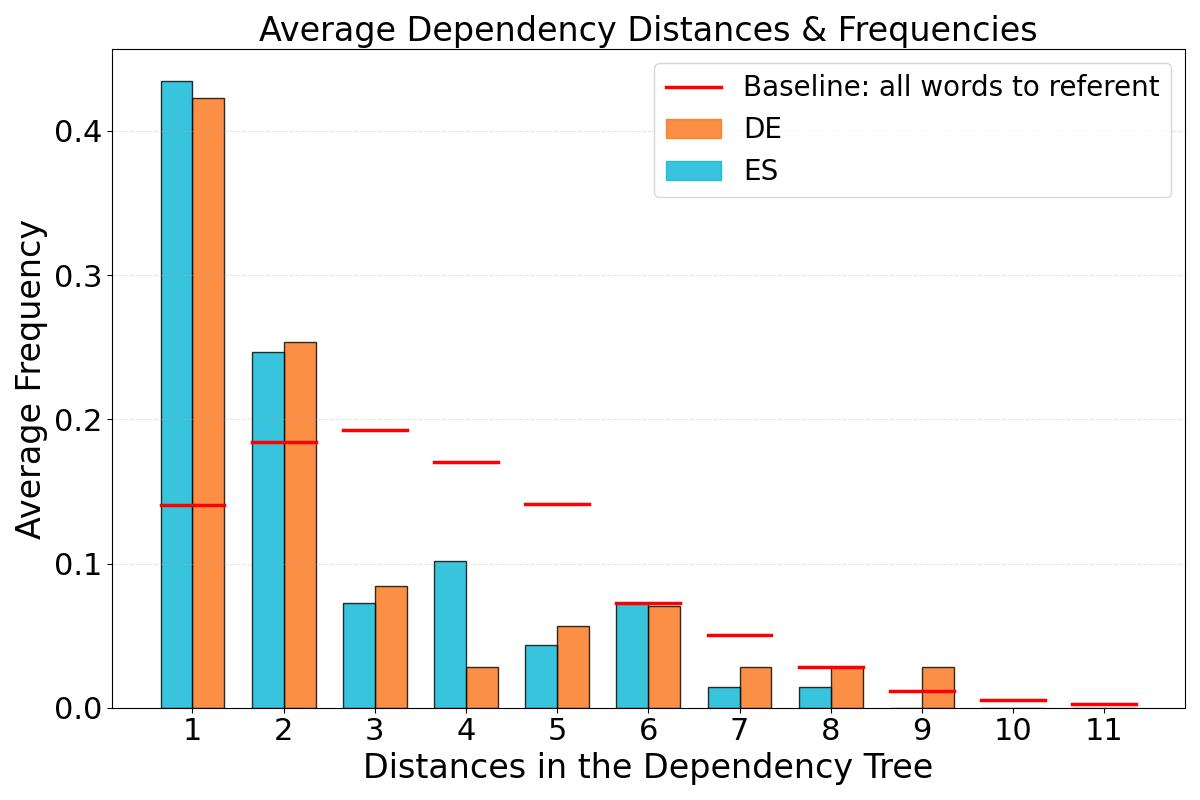}
  \caption {Average dependency distance of salient words to target referent (Approach 4), in comparison to the distribution in the source data.}
  \label{fig:dependency}
\end{figure}

\paragraph{Dependency Distance}
For both EN$\rightarrow$DE and EN$\rightarrow$ES, Figure \ref{fig:dependency} shows the dependency distances measured between salient words and target referents defined as the number of syntactic links (edges) separating the two nodes in a given dependency tree. These distances show how close salient words (that influence an MT's translation decision in terms of gender) are to the target referent in question, grammatically. Here again, the general trends of dependency distances for context words considered salient are similar for both language pairs. The figure clearly depicts that most salient source words are at a distance of 1 away from the target referent, with a frequency of 43\% and 42\% for ES and DE, respectively. This is followed by salient source words at a distance of 2 away from the target referent, with a frequency of 25\% for both. All following distances are negligible, with the biggest difference in terms of language being at distance 4. The red horizontal lines in Figure \ref{fig:dependency} represent the average frequency of dependency distance between the target referent and all other words in the sentence. Interestingly, words at a dependency distance of 1 (and to a lesser extent also 2) are considered salient much more frequently in contrast to their distribution in the overall data. Distances longer than 2 are considerably less salient than their overall frequency in the data.

In comparison, in \citet[p. 230]{hackenbuchner-etal-2025-clin} it has been shown that the context words that most affected human perceptions of gender were at a dependency distance of 3, 1, 4, 2, 5 and 6 (in that order), with the distributions of these distances being similar to each other. Our findings align with this pattern, as reflected in the similarly distributed dependency distances between the target word and all other words in each sentence (i.e. the red lines depicted in Figure \ref{fig:dependency}). These results show that, for the translation of gender, the model is predominantly influenced by words that are very close in dependency structure, while humans are influenced by broader context including words further away from the target referent in question.

\subsection{Qualitative Analysis of Outliers}
The main analysis is conducted on the relative top 15\% of salient source words, where an average overlap with human annotations achieves 0.851. However, this leaves $\sim$15\% of source words that have been considered salient by the MT model but have not been annotated by humans. There does not seem to be a clear pattern among these `outliers'. Verbs were the most common type of word, comprising 37.5\% of these outliers, which were considered salient by a model but not annotated. Apart from that, unlike humans, the model considered salient numbers as well as two words that included an <unk> in the token (e.g., \textit{Who<unk>s}), where an apostrophe was not properly recognised. Furthermore, in one sentence, the system considered salient a name, `Oreskovich', that humans did not annotate. These last two scenarios could be explained by models being more influenced by unique tokens.

\section{Discussion}\label{discussion}
In this exploratory paper, we show that contextual cues (source words) that have been shown to influence human perception of gender are also found among the tokens that influence a model’s translation of gender in the target language (i.e. that increase the probability of a certain gender output over another). This shows that humans and models are influenced by relatively similar contexts regarding gender, even in ambiguous scenarios. However, the most frequent tokens that influence an MT system seem to linguistically differ from human annotations both in terms of their POS and grammatical structure.

Regarding our approaches to analyse saliency of input tokens from different perspectives, interestingly, approaches 1 (overall top 5\%) and 2 (top word) reached the same highest model-human overlap. This is due to the fact that the top one word is often the only word included in the top overall 5\% of salient words, particularly as the average sentence length is only $\sim$19 words. Approach 3 likely had a lowest model-human overlap as normalised values prevent direct comparison across intra-sentential tokens, while it does allow for comparison across sentences. Future work could include quantifying model-human comparison by different means, such as using Plausibility Evaluation of Context Reliance (PECORE) to detect context-sensitive tokens and their disambiguating rationales in MT \citep{sarti-etal-2024}. Furthermore, the fact that there is a lower model-human overlap when taking only annotations into consideration where at least two annotators agree (see Figure \ref{fig:approach4_annotator_comparison}) relates back to human perceptions of gender varying extremely among annotators, with an overall ‘fair’ IAA ($\kappa$~=~0.364, Fleiss' Kappa).

Previous research on detecting (salient) input tokens related to anaphora resolution shows that for mistranslations, pronouns were not salient input tokens and were thus overlooked by the model \citep{attanasio-etal-2023-tale} and that default masculine pronouns predominantly receive weaker attention, while feminine pronouns elicit more localised activations, requiring more attention \citep{manna-etal-2025-paying}.
Even though we cannot directly compare our findings, as we did not look at anaphora resolution, we found that pronouns referring to other target referents in the data only comprise 9.7/7.5\% (DE/ES) of all salient words. \citet{yin-neubig-2022-interpreting} show that a model is correctly influenced by gendered proper nouns and pronouns in the source when determining the gender of pronouns, nouns or determiners. Similarly, however, their model is incorrectly influenced by proper nouns and pronouns of a different gender in the input when wrongly predicting the gender of a pronoun. In the study presented here, we cannot define whether a model `correctly' or `incorrectly' predicts the gender of a target referent, as we are looking at ambiguous scenarios. However, we show that proper nouns and pronouns are the third and fourth most influential POS, surpassed by nouns and verbs, which are demonstratively more influential in the prediction of gender.

Despite a high model-human overlap of 0.851, our results show differences in POS and dependency distances for human annotations and salient words. Noteworthy is that the MT is overarchingly influenced by two POS categories (NOUNs and VERBs), while POS categories annotated by humans had a more balanced distribution. This could partially stem from the fact that humans annotated individual words but also phrases (split into words for this analysis). This similarly could have an effect on the imbalance in the model-human overlap of dependency distances. Phrases encompass different POS categories and dependency distances (to the target referent). Looking back at Figure \ref{fig:fig1}, the words in the annotated phrase ``making fun of me'' encompass four different POS categories (VERB, NOUN, ADP, PRON) and three different dependency distances (3--5). Annotated phrases are therefore likely to be a primary factor leading to model-human differences in POS categories and dependency distances.

Something to take into consideration is the fact this this work was tested on the \texttt{OPUS-MT} model, which is a widely-tested open-source model, but cannot keep up with state-of-the-art MT systems. The translation quality will therefore not compare to SOTA models. However, this work analyses translation \textit{differences} in terms of gender, not translation quality. For this task, the selected model is perfectly suited to assess this phenomena, as outlined in Section \ref{methodology}.


\section{Conclusion}\label{conclusion}
In this paper, we leverage an interpretability technique (i.e. contrastive analysis) to examine gender choices in machine translation. Applying an attribution-based analysis on contrastive translations, we evaluated which source context influences (triggers) the gender translation of a target referent in an ambiguous sentence. We find that (i) overall, the most salient source words to the given MT model (\texttt{OPUS MT}) highly overlap (0.851) with words that trigger human perceptions of gender, and that (i) nouns, verbs, and adjectives (and, lesser, pronouns, proper nouns and numbers) affect the MT model's choice of gender the most. Our findings highlight that, to a large extent, our model and humans are influenced by the same words in (an ambiguous) context. Despite a high overlap, we show persistent model-human differences in POS categories and dependency distances. Our contribution moves away from simply \textit{measuring} bias to \textit{exploring} its origins. We find it crucial to improve our understanding of model behaviour by shining a light on \textit{how} an MT model handles gender, and how this compares to human perceptions of gender. We hope that our study leads to increased research in this domain, with the goal of leading to better solutions.

\section{Ethical Considerations}
In this paper, we examine model translation bias in a gender-ambiguous context, based on input saliency attribution attained by means of contrastive explanations. We do not claim to provide a \textit{faithful explanation} \citep{jacovi-goldberg-2020,bastings-etal-2022} on bias analysis in MT models. Rather, we demonstrate that for a better understanding of how models exhibit gender bias, attribution methods could and should be used for exploring social biases in sequence-to-sequence models. We assume a simplified concept of binary gender to allow for a more straightforward evaluation of the results, as current MT models continue to struggle with adequately translating inclusive and non-binary gender \citep{hackenbuchner-etal-2025-genderous,pranav-etal-2025}. We address gender bias in terms of the broader issue of representational harm \cite{blodgett-etal-2020}, categorised into two types: under-representation and stereotyping.

\section{Limitations}\label{limitations}

The dataset comprising 60 sentences used for this study is very small in today's field of NLP research. However, this was an experimental study on the interplay of these fields, for which the pre-existing natural ambiguous dataset and annotations by 20 annotators of different genders served useful. In future work, we aim to extend the analysis based on contrastive translations to a larger dataset (once available).

Another limitation of this study is that we tested only one model (\texttt{OPUS-MT}) and two language pairs (EN$\rightarrow$DE/ES). Future work could compare different models and languages. However, the number of open-source models that can currently be used for a contrastive approach, as taken in this study, is limited. The limitation thus not only stems from the study design from this paper but is a general limitation of the methodology and toolkits available. Model tokenization has an impact on how input tokens influence and are considered salient for a model's translation decision. Subwords are merged in this paper and scores summed. Tokenization varies by models, another reason why we focussed on one model in this study, and summing subwords and comparing words across different models could be faulty.

As touched upon in the body of the paper, there is great annotation diversity among the annotators with a low IAA. This shows that human perception of gender cannot be taken as default. Furthermore, in this study, we focus on a simplified binary gender by contrasting masculine and feminine translation outputs. We apply this approach as our chosen MT model does not provide gender-inclusive or gender-neutral translations (it could, theoretically, provide traditional German neutral referent translations).

\section{Acknowledgements}
We would like to thank the anonymous reviewers for their invaluable feedback. This study is part of a strategic basic PhD research (1SH5V24N) fully funded by The Research Foundation – Flanders (FWO) for the time span of four years, from 01.11.2023 until 31.10.2027, and hosted within the Language and Translation Technology Team (LT3) at Ghent University.

\section{Bibliographical References}\label{reference}

\bibliographystyle{lrec2026-natbib}
\bibliography{lrec2026}

\appendix

\section{Micro vs. Macro Precision Scores}\label{app:macro-micro}

\begin{table}[h]
\begin{tabular}{cc|c|c|c|c|c|c|c|c|c|c}
\hline
\hline
\multicolumn{2}{l}{\textbf{1 - Top \%}} & \textbf{0.05} & \textbf{0.10} & \textbf{0.15} & \textbf{0.20} & \textbf{0.25} \\
\hline
\multirow{2}{*}{\textbf{DE}} & $P_{mi}$  & 0.817 & 0.795 & 0.777 & 0.766 & 0.752 \\
                             & $P_{ma}$ & 0.817 & 0.811 & 0.804 & 0.780 & 0.775 \\
\hline
\multirow{2}{*}{\textbf{ES}} & $P_{mi}$ & 0.879 & 0.838 & 0.800 & 0.797 & 0.768 \\
                             & $P_{ma}$ & 0.850 & 0.819 & 0.792 & 0.791 & 0.767 \\
\hline
\hline
\multicolumn{2}{l}{\textbf{2 - Top Word}}    \\
\hline
\multirow{2}{*}{\textbf{DE}} & $P_{mi}$  & 0.817 \\
                             & $P_{ma}$ & 0.817 \\
\hline
\multirow{2}{*}{\textbf{ES}} & $P_{mi}$ & 0.879 \\
                             & $P_{ma}$ & 0.850 \\
\hline
\hline
\multicolumn{2}{l}{\textbf{3 - Min. Attr.}} & \textbf{0.01} & \textbf{0.02} & \textbf{0.03} & \textbf{0.04} & \textbf{0.05} & \textbf{0.06} & \textbf{0.07} & \textbf{0.08} & \textbf{0.09} & \textbf{0.10} \\
\hline
\multirow{2}{*}{\textbf{DE}} & $P_{mi}$  & 0.533 & 0.608 & 0.657 & 0.711 & 0.736 & 0.753 & 0.783 & 0.784 & 0.779 & 0.791 \\
                             & $P_{ma}$ & 0.562 & 0.625 & 0.669 & 0.741 & 0.780 & 0.787 & 0.784 & 0.729 & 0.642 & 0.639 \\
\hline
\multirow{2}{*}{\textbf{ES}} & $P_{mi}$ & 0.522 & 0.599 & 0.652 & 0.714 & 0.752 & 0.765 & 0.797 & 0.814 &                                               0.812 & 0.831 \\
                             & $P_{ma}$ & 0.537 & 0.599 & 0.645 & 0.697 & 0.746 & 0.771 & 0.801 & 0.731 &                  0.674 & 0.625 \\
\hline
\hline
\multicolumn{2}{l}{\textbf{4 - Rel. Attr.}} & \textbf{0.05} & \textbf{0.10} & \textbf{0.15} & \textbf{0.20} & \textbf{0.25} & \textbf{0.30} & \textbf{0.35} & \textbf{0.40} & \textbf{0.45} & \textbf{0.50} \\
\hline
\multirow{2}{*}{\textbf{DE}} & $P_{mi}$ & 0.817 & 0.817 & 0.806 & 0.816 & 0.792 & 0.782 & 0.766 & 0.769 & 0.756 & 0.732 \\
                             & $P_{ma}$ & 0.817 & 0.817 & 0.808 & 0.833 & 0.817 & 0.806 & 0.790 & 0.795 & 0.770 & 0.753 \\
\hline
\multirow{2}{*}{\textbf{ES}} & $P_{mi}$ & 0.879 & 0.881 & 0.897 & 0.841 & 0.838 & 0.819 & 0.787 & 0.770 &                                   0.768 & 0.741 \\
                             & $P_{ma}$ & 0.850 & 0.850 & 0.867 & 0.828 & 0.819 & 0.808 & 0.772 & 0.756 & 0.758 & 0.732 \\
\hline                                             
\end{tabular}
\begin{minipage}{1.9\linewidth}
\caption{\label{tab_app:approaches}
    Micro versus macro precision scores across the four threshold approaches, for all annotators for DE and ES. Approach 1: Top Percentage (top overall percentage: 5\%, 10\%, 15\%, 20\%, 25\%), Approach 2: Top Word, Approach 3: Minimum Attribution Scores (min. saliency score per word of 0.01 etc.), Approach 4: Relative Attribution Scores (top percent relative per sentence: top 5\%, 10\%, 15\%, 20\%, 25\%, 30\%, 35\%, 40\%, 45\%, 50\%.). Macro precision is the arithmetic mean, obtained by calculating individual precisions for each sentence and then taking an overall average of the precisions across all sentences.}
\end{minipage}
\end{table}

\section{Macro Precision Scores}\label{app:macro_scores}

\begin{figure}[h]
\centering
  \includegraphics[width=0.85\columnwidth]{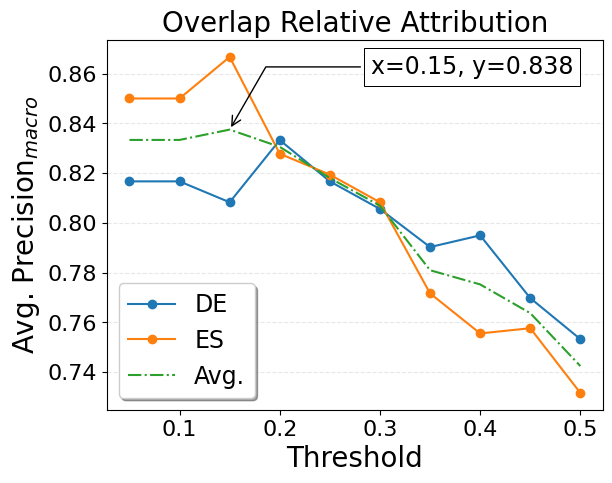}
  \caption {Comparative scoring of overlapping source words for Approach 4, depicting average macro precision scores for DE, ES and their average. The x-axis should be interpreted in percent (e.g., 10\%).}
  \label{fig_app:approach4}
\end{figure}

\clearpage
\section{Instructions for Annotators}\label{app:annotation_instructions}

\begin{figure}[h!]
  \includegraphics[width=1.9\linewidth]{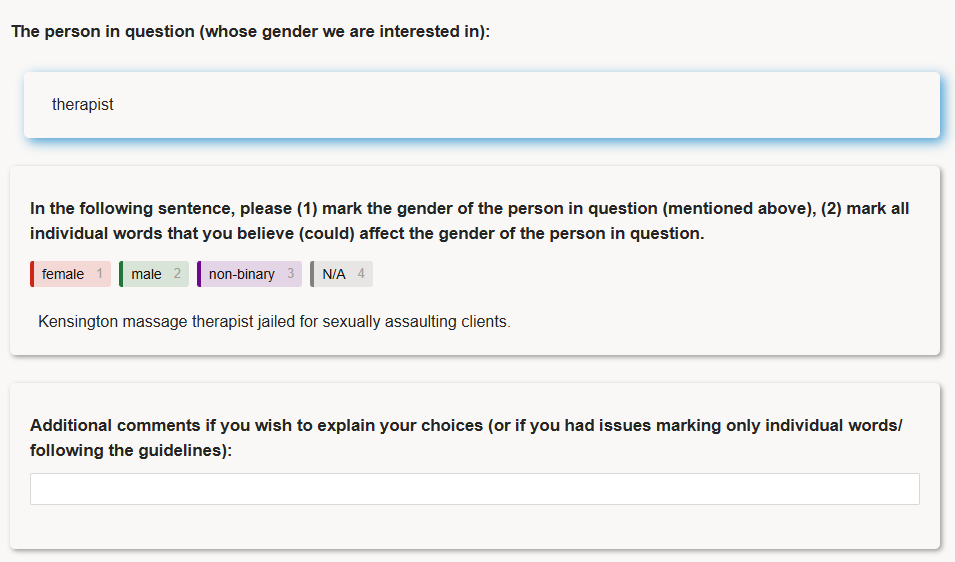} \hfill
  \begin{minipage}{1.9\linewidth}
  \caption {Instructions that annotators were given, taken from \citet{hackenbuchner-etal-2025-clin}.}
  \label{fig:annotation_example}
  \end{minipage}
\end{figure}

\end{document}